\def\year{2020}\relax
\documentclass[letterpaper]{article} 
\usepackage{aaai20}  
\usepackage{times}  
\usepackage{helvet} 
\usepackage{courier}  
\usepackage[hyphens]{url}  
\usepackage{graphicx} 
\usepackage{graphicx}  
\title{Towards Action Model Learning for Player Modeling}
\author{Abhijeet Krishnan, Aaron Williams, Chris Martens\\
POEM Lab,\ North Carolina State University\\
Raleigh, NC, 27606\\
\{akrish13,\ amwill19\}@ncsu.edu, martens@csc.ncsu.edu \\ 
}

\usepackage{subcaption}
\usepackage{listings}
\usepackage{array}
\usepackage[ruled]{algorithm2e}
\usepackage{amsmath, centernot}
\usepackage{amssymb}
\usepackage{multirow}

\newcommand\citeA[1]{\citeauthor{#1} (\citeyear{#1})}

\lstdefinelanguage{pddl}
{
  sensitive=false,    
  morecomment=[l]{;}, 
  alsoletter={:,-},   
  morekeywords={
    define,domain,problem,not,and,or,when,forall,exists,either,
    :domain,:extends,:requirements,:types,:objects,:constants,
    :predicates,:action,:parameters,:precondition,:effect,:functions,
    :fluents,:primary-effect,:side-effect,:init,:goal,assign
    :strips,:adl,:equality,:typing,:conditional-effects, :metric, minimize,
    :negative-preconditions,:disjunctive-preconditions,
    :existential-preconditions,:universal-preconditions
  },
  keywords=[2]{}, 
  keywords=[3]{clear, at, at-goal, IS-GOAL, IS-NONGOAL, MOVE-DIR, on, smaller, empty, neighbor}, 
  keywords=[4]{move, push-to-nongoal, push-to-goal}, 
  keywords=[5]{} 
}

\begin{document}

\maketitle

\begin{abstract}
Player modeling attempts to create a computational model which accurately approximates a player's behavior in a game. 
Most player modeling techniques rely on domain knowledge and are not transferable across games. 
Additionally, player models do not currently yield any explanatory insight about a player's cognitive processes, such as the creation and refinement of mental models.
In this paper, we present our findings with using {\em action model learning} (AML), in which an action model is learned given data in the form of a play trace, to learn a player model in a domain-agnostic manner. 
We demonstrate the utility of this model by introducing a technique to quantitatively estimate how well a player understands the mechanics of a game.
We evaluate an existing AML algorithm (FAMA) for player modeling and develop a novel algorithm called Blackout that is inspired by player cognition. 
We compare Blackout with FAMA using the puzzle game Sokoban and show that Blackout generates better player models.
\end{abstract}

\section{Introduction}

Player modeling is the study of computational models of players in games \cite{yannakakis2013player}. 
It sees varied and widespread usage in today's video game industry. 
Imangi Studios, developers of the popular \textit{Temple Run} (Imangi Studios 2011) series of mobile games, collect player telemetry in order to analyse player behavior and provide customized gameplay experiences.
\textit{Forza Motorsport 5} (Turn 10 Studios 2013) implements a \textit{Drivatar} system which learns to mimic the player's behavior in the game and simulate the player in other races.

Despite the multitude of techniques used to build player models, such as self-organizing maps \cite{drachen2009}, Bayesian networks \cite{yannakakis2005}, and multi-layer perceptrons \cite{pedersen2010content}, many of them rely on features extracted from domain knowledge of the game's rules and as such cannot be generalized easily, except perhaps to games of the same genre. 
The inability to easily train new models for different games using the same technique presents a barrier to any single technique's adoption.

Furthermore, while current techniques aim to {\em predict} player actions, we argue that there is a corresponding need to be {\em explain} their underlying cognitive processes. There is empirical evidence in cognitive science that game players build {\em mental models}, knowledge structures capable of simulating the system they are interacting with and predicting and explaining the outcomes of scenarios, while playing games \cite{boyan2018model}. 
Mental models may start out erroneous, but improve over time as the player learns from which of their attempted actions succeed and fail. We believe that mental model alignment is one of the most important aspects of using games for impactful applications, such as education and training, and that player modeling techniques should be designed to yield insight into this cognitive process.

To address these needs, we propose {\em action model learning} (AML), in which an action model is learned from play traces, as a viable technique.
Action models can be used to learn player models in any game which can be represented in a planning formalism like PDDL. 
There is a rich body of literature on learning action models from action traces which we can leverage to learn action models. 

In this paper, we describe our efforts to build an action model-based player model which can be used to characterize player competency.
To ascertain the feasibility of action model learning as an approach to player modeling, we started with a modern algorithm known as FAMA~\cite{aineto2019104} out-of-the-box. In parallel, we developed an in-house alternative that we call Blackout, motivated by mental model alignment and taking failed actions into account. 
We test both approaches on the puzzle game Sokoban and evaluate their output, finding that Blackout outperforms FAMA for the task of player modeling.
We discuss both techniques' advantages and limitations and suggest avenues for future work.

While this paper's quantitative results focus on the comparison between AML and Blackout, more generally, we argue that AML is a {\em tractable} and {\em domain-agnostic} approach to player modeling, and that the learned action model is a {\em useful} player model. We justify the tractability claim by successfully applying both FAMA and Blackout to the task of learning a player model from action traces, measuring their efficiency on problems of various sizes as well as precision and recall. We justify the domain-agnosticity claim by successfully learning action models for two additional domains: Hanoi (the Tower of Hanoi puzzle) and N-puzzle (a sliding tile puzzle), using publicly available domain files\footnote{\url{http://planning.domains/}} and trajectories from FAMA's evaluation dataset. 
We justify the usefulness claim by presenting a technique to quantify a player's mechanical mastery of a game given their action model-based player model.

\section{Related Work}

Player modeling is a relatively new field, previously studied under the umbrella of HCI in the form of user modeling 
~\cite{biswas2018user} 
and student modeling.
~\cite{chrysafiadi2013student} 
Recent surveys on player modeling provide useful taxonomies~ \cite{smith2011inclusive}, an overview of techniques that have been used for player modeling generally \cite{machado2011player} or for MMORPGs \cite{harrison2011using}, and challenges commonly encountered in the field, including the reliance on knowledge engineering \cite{hooshyar2018datadriven} that motivates our work.

Domain-agnostic approaches to player modeling have been attempted before.
\citeA{snodgrass2019peas} describe a general player model using the PEAS framework, which presents a theoretical framework for providing game recommendations, but does not account for in-game player behavior.
\citeA{nogueira2014fuzzy} use physiological readings of players to model their emotional response to game events.
While input-agnostic, this technique relies on physiological data from sensors, which is difficult to acquire.
Our approach requires only gameplay traces, which can be easily done by making minor modifications to the game engine code.
Deep neural networks have been used to simulate the actions of human players in interactive fiction games \cite{pengcheng2018fidelity} and MMORPGs \cite{pfau2018dpbm} and to generate new levels in a platformer game based on learned player preferences \cite{summerville2016learning}.
These techniques rely only on easily obtainable input like gameplay logs or video recordings, and do not use any knowledge engineering to train the model.
We believe our approach has two advantages: that it does not require as much data to learn, and, by generating a rule-based model, offers more explanatory power for player behavior.

To the best of our knowledge, this is the first-known application of AML to player modeling.
The literature on AML has primarily focused on learning sound action models for use by an automated planner, with little attention paid to modeling the player's cognitive processes. 
\textit{Human-aware planning} attempts to create planning agents which can take human mental models into account while planning \cite{chakraborti2017cognitive}, while we attempt to treat a human as a planning agent and learn a domain model which mirrors their mental model.
\citeA{serafini2019incremental} introduce a perception function which maps sensor data to state variables in order to learn action models. 
We believe investigating player perception functions for cases like player disability to learn action models would be fruitful to pursue.
An early AML algorithm called OBSERVER uses action failures to learn action preconditions
~\cite{wang1995learning}, which mirrors our approach.

\section{Background}

Action model learning (AML) is situated within the tradition of automated planning, in which an {\em action model} describes the preconditions and effects of each action an agent may take in a world. Preconditions and effects are propositional formulas over fluents (predicates whose truth changes with time). A {\em plan} is a sequence of actions, each instantiated with terms for each parameter position, such that the effects of each action entail the preconditions of the action following it.
AML, then, is the problem of discovering an action model from a set of observed plans.

More formally, an AML algorithm takes as input a number of \textit{action traces}, each a sequence $\tau = \langle s_0, a_1, s_1, a_2, s_2, \cdots, a_n, s_n \rangle$, where $s_i$ are states and $a_i$ are actions, and returns as output an action model, which is a specification of the preconditions and effects of every action in the domain. These action traces are usually assumed to be fully observed, i.e. every fluent of a state is present, but various developments in AML algorithms allow action models to be learned from action traces with partially observed or noisy states as well.
An action model is represented as a STRIPS-style domain, with the most common representation format being PDDL. 

The primary domain used in development and testing is the classic Japanese block-pushing puzzle game Sokoban, which we use to demonstrate the feasibility of our approach.
We select it based on its relative minimalism and existing formalizations in PDDL as part of the International Planning Competition (IPC) benchmark suite \cite{ipc2011survey}.
In Sokoban, there are stones (or crates) scattered around a warehouse floor, which the player must push to certain specified locations around the warehouse, but can only push the stones in cardinal directions, cannot climb over the stones or other obstacles, and can only push one stone at a time. For an overview of Sokoban's rules and its characteristics as a search problem, we refer readers to \citeA{junghanns1997sokoban}.
The specific representation of Sokoban we use is a domain that appeared in the 2011 International Planning Competition,\footnote{\url{https://github.com/potassco/pddl-instances/blob/master/ipc-2011/domains/sokoban-sequential-satisficing/domain.pddl}} modified to remove action costs.

\section{Player Modeling with FAMA}

\begin{figure}[ht]
    \centering
    \begin{subfigure}[t]{0.45\columnwidth}
        \centering
        \begin{lstlisting}[language=PDDL]
(:action move
	:parameters 
    (?p - player 
     ?from - location 
     ?to - location 
     ?dir - direction)
	:precondition (and 
		(clear ?to) 
    (at ?p ?from)
		(is-nongoal ?from))
	:effect (and 
		(clear ?from)
    (at ?p ?to)
    (move-dir ?from ?to ?dir)
		(not (at ?p ?from))
		(not (clear ?to))))\end{lstlisting}
    \end{subfigure}\hfill
    \begin{subfigure}[t]{0.45\columnwidth}
        \centering
        \begin{lstlisting}[language=PDDL]
(:action move
	:parameters 
    (?p - player 
     ?from - location 
     ?to - location 
     ?dir - direction)
	:precondition (and
		(clear ?to) 
		(at ?p ?from)
    (move-dir ?from ?to ?dir))
	:effect (and 
		(clear ?from)
    (at ?p ?to)
		(is-nongoal ?to)
		(is-nongoal ?from)
		(not (at ?p ?from))
		(not (clear ?to))))\end{lstlisting}
    \end{subfigure}
    \caption{A comparison between the \texttt{move} action learned by $M_1$ and $M_2$}
    \label{fig:model-comparison}
\end{figure}

AML attempts to learn the domain model of a game. 
In applying AML to player modeling, we treat the player's mental model of the game's rules as a domain, and attempt to learn it using an AML algorithm.
Our goal is to learn an action model corresponding to the player's mental model of the game's mechanics.

To determine baseline feasibility of AML as an approach to player modeling, we started with an out-of-the-box algorithm known as FAMA~\cite{aineto2019104}.
FAMA performs AML by compiling the learning task into a classical planning task.
This is done by creating actions in the compiled planning domain which correspond to inserting fluents in the preconditions or effects of various actions in the planning domain to be learned.
The solution plan is thus a sequence of actions which builds up the output domain model and verifies its consistency. 

FAMA has properties which are useful for our problem. 
It is capable of working with action traces having incomplete states (states with missing fluents) and missing actions. 
Since it utilizes classical planning, we can use a variety of existing planners in our system based on their properties such as computation time and memory availability. 
The action models it produces are formally sound and can be used to generate new trajectories.
It is capable of learning action models from multiple different trajectories.

In lieu of a human player, we use the FastDownward 19.12 planning system to generate a solution path for each of two custom levels, then convert each solution into an action trace which is input to FAMA.
We use FAMA with full state observability to successfully learn two action models $M_1$ and $M_2$ from each of the two obtained action traces. 
Figure \ref{fig:model-comparison} shows the |move| action learned by both models.

\newcommand{\systemname}[0]{Blackout }

\section{Player Modeling with \systemname}

Our new AML algorithm, Blackout, takes advantage of stronger assumptions that can be made of the input (such as full observability) to improve execution speed and model accuracy.
Blackout also takes action {\em failures} into account based on the cognitive theory that failed actions inform players' mental models.

The inputs to \systemname are (1) a play trace and (2) a domain file $D = \{\Theta, \Psi, \alpha\}$, where $\Theta$ is the set of types, $\Psi$ is the set of predicates, and $\alpha$ is the action model.
The action model consists of actions with positive and negative preconditions and effects represented by $\texttt{pre}^{+}(a)$,$\texttt{pre}^{-}(a)$,$\texttt{eff}^{+}(a)$ and $\texttt{eff}^{-}(a)$ respectively. 
We assume deterministic action effects with full observability of states and actions and no noise.
We believe this is a reasonable assumption given how the trajectories are obtained.
\systemname outputs a STRIPS action model which is not guaranteed to be sound.

\systemname operates in three steps: first, it produces an initial action model by analyzing the differences between the \textit{pre-state} and \textit{post-state} for every action in the trajectory.
Next, it uses information from failed action executions to improve the initial action model.
Finally, it computes invariants which hold for predicates in the domain to further improve the action model. We now explain each step in detail.

\subsection{Step 1: Successful Action Analysis}

We first compute the effects of actions by calculating the \textit{delta state} for each action execution. 
For discovering preconditions, we first calculate the set of all possible fluents which could be valid preconditions given the domain and the action's bindings.
This is done by binding all possible combinations of objects already bound to an action's arguments, given by \texttt{obj}$(a)$, to all predicates in the domain.
Assignment of a set of objects $\omega$ to a predicate $p$ is represented by $p(\omega)$, assuming $\|\omega\| = \texttt{arity}(p)$ and there is no type mismatch.
Applicable preconditions are first \textit{generalized} in the function $G$ i.e. their bindings are replaced with variables and then added to the precondition lists. 
The action model output in this step may contain superfluous and erroneous preconditions.

\begin{algorithm}[ht]
    \SetAlgoLined
    \SetKwFunction{obj}{obj}
    \SetKwFunction{ar}{arity}
    \SetKwFunction{effp}{$\text{eff}^{+}$}
    \SetKwFunction{effm}{$\text{eff}^{-}$}
    \SetKwFunction{prep}{$\text{pre}^{+}$}
    \SetKwFunction{prem}{$\text{pre}^{-}$}
    \SetKwFunction{generalized}{$G$}
    \KwData{a set $T$ of trajectories of the form $\langle s_0, a_0, s_1, a_1, s_2, \cdots, a_{n-1}, s_n \rangle$, a reference domain model $D = \{\Theta, \Psi, \alpha\}$}
    \KwResult{a STRIPS action model $A$}
    
    \Begin{
        Initialize empty action model $A$\;
        \ForEach{trajectory $t \in T$}{
            \ForEach{$\langle s - a - s' \rangle$ in $t$}{
                $\overline{s} \leftarrow s \setminus \{p \mid p \in s, \obj{p} \nsubseteq \obj{a} \}$\;
                $\overline{s}' \leftarrow s' \setminus \{p \mid p \in s', \obj{p} \nsubseteq \obj{a} \}$\;
                $F \leftarrow \{p(\omega) \mid p \in \Psi, \omega \in \obj{a}^{\ar{p}}\}$\;
                $\prep{a} \leftarrow \generalized{$F \setminus \overline{s}$}$\;
                $\prem{a} \leftarrow \generalized{$F \setminus \overline{s}'$}$\;
                $\effp{a} \leftarrow \generalized{$\overline{s}' \setminus \overline{s}$}$\;
                $\effm{a} \leftarrow \generalized{$\overline{s} \setminus \overline{s}'$}$\;
            }
        }
        \Return A
    }
    \caption{Successful Action Analysis}
    \label{fig:blackout-1}
\end{algorithm}

\subsection{Step 2: Failed Action Analysis}

Failed actions are actions which the player attempts to perform in-game, but which cannot be completed due to the preconditions of an action not being met. 
It indicates a mismatch between the player's mental model of the game's mechanics and the actual action model representing the game. 

\systemname makes the assumption that players notice the failure of action execution and update their mental model to account for this. 
We record failed actions in a manner similar to successful actions in the trajectory, with failed actions forming triplets of the form $\langle s - a_f - s \rangle$. 
The pre-state and post-state are identical since the action failed to execute. 
\systemname attempts to identify the predicates in the pre-state which caused the action $a_f$ to fail to execute due to violating its preconditions.
These are stored in the sets $R_{a_f}^{+}$ and $R_{a_f}^{-}$ for positive and negative preconditions respectively. 

The detailed algorithm is described in Algorithm \ref{fig:blackout-2}. 

\begin{algorithm}[ht]
    \SetAlgoLined
    \SetKwFunction{obj}{obj}
    \SetKwFunction{ar}{ar}
    \SetKwFunction{effp}{$\text{eff}^{+}$}
    \SetKwFunction{effm}{$\text{eff}^{-}$}
    \SetKwFunction{prep}{$\text{pre}^{+}$}
    \SetKwFunction{prem}{$\text{pre}^{-}$}
    \SetKwFunction{generalized}{$G$}
    \KwData{a set $T$ of trajectories of the form $\langle s_0, a_0, s_1, a_1, s_2, \cdots, a_{n-1}, s_n \rangle$ with failed actions $a_f$, a reference domain model $D = \{\Theta, \Psi, A\}$, action model $A$ from step 1}
    \KwResult{a STRIPS action model $A$}
    
    \Begin{
        \ForEach{trajectory $t \in T$}{
            \ForEach{$\langle s - a_f - s \rangle$ in $t$}{
                $R_{a_{f}}^{+} \leftarrow R_{a_{f}}^{+} \cup (\prep{$a_f$} \setminus s)$\;
                $R_{a_{f}}^{-} \leftarrow R_{a_{f}}^{-} \cup (\prem{$a_f$} \cap s)$\;
                \uIf{$R_{a_{f}}^{+}$ is a singleton set $\And$ $R_{a_{f}}^{-} = \emptyset$}{
                    Mark $p \in R_{a_{f}}^{+}$ as confirmed\;
                }
                \uElseIf{$R_{a_{f}}^{+} = \emptyset$ $\And$ $R_{a_{f}}^{-}$ is a singleton set}{
                    Mark $p \in R_{a_{f}}^{-}$ as confirmed\;
                }
                \Else{
                    Mark $\langle R_{a_{f}}^{+}, R_{a_{f}}^{-}, a_f \rangle$ as ambiguous\;
                }
            }
        }
        \ForEach{action $a \in A$}{
            \prep{$a$} $\leftarrow G(\{p \mid p \in \prep{$a$}$ where $p$ is confirmed\})\;
            \prem{$a$} $\leftarrow G(\{p \mid p \in \prem{$a$}$ where $p$ is confirmed\})\;
        }
        \Return A
    }
    \caption{Failed Action Analysis}
    \label{fig:blackout-2}
\end{algorithm}

\subsection{Step 3: Invariant Extraction}

In this step, \systemname identifies invariants that hold among pairs of predicates in the domain.
An invariant between two predicates $p$ and $q$ for some shared variable $v$ is a relation on $\{(p, q), (p, \lnot q), (\lnot p, q), (\lnot p, \lnot q)\}\} \times \mathbb{B}$.
It describes what combinations of $p$ and $q$ are permitted to be attached to $v$ or not (attachment denoted by $p$ and $q$, lack thereof by $\lnot p$ and $\lnot q$) after an action has been performed on $v$.
\systemname uses these invariants to resolve any ambiguities regarding the predicates responsible for action failure in step 2, and in doing so further refine the action preconditions.

\begin{algorithm}[]
    \SetAlgoLined
    \SetKwFunction{obj}{obj}
    \SetKwFunction{ar}{ar}
    \SetKwFunction{effp}{$\text{eff}^{+}$}
    \SetKwFunction{effm}{$\text{eff}^{-}$}
    \SetKwFunction{prep}{$\text{pre}^{+}$}
    \SetKwFunction{prem}{$\text{pre}^{-}$}
    \SetKwFunction{generalized}{$G$}
    \SetKwFunction{var}{var}
    \SetKwFunction{primitive}{PrimitiveRule}
    \SetKwFunction{invariant}{Invariant}
    \SetKwFunction{merge}{Merge}
    \SetKwFunction{type}{Type}
    \KwData{a set $T$ of trajectories of the form $\langle s_0, a_0, s_1, a_1, s_2, \cdots, a_{n-1}, s_n \rangle$ with failed actions $a_f$, a reference domain model $D = \{\Theta, \Psi, \alpha\}$, action model $A$ from step 2, $R_{a_{f}}^{+}$ and $R_{a_{f}}^{-}$ from step 2}
    \KwResult{a STRIPS action model $A$}
    
    \Begin{
        $P \leftarrow \emptyset$\;
        \ForEach{action $a \in A$}{
            \ForEach{$ (p_1, p_2) \in ( $ \effp{$a$} $\cup $ \effm{$a$}$ )^2, p_1 \neq p_2$}{
                \ForEach{variable $v \in$ \var{$p_1$} $\cap$ \var{$p_2$}}{
                    $P \leftarrow P \cup$ \primitive{$p_1, p_2, v$}\;
                }
            }
        }
        \ForEach{primitive rule $r(p, q, v) \in P$}{
            \If{$ \exists a \in A, q \in $ \effp{$a$}$ \cup $ \effm{$a$}$ , p \notin $ \effp{$a$} $ \cup $ \effm{$a$} }{
                $P \leftarrow P \setminus \{r\}$\;
            }
        }
        $I \leftarrow \emptyset$\;
        \ForEach{primitive rule $r(p, q, v) \in P$}{
            $M \leftarrow \{ r \}$\;
            \ForEach{primitive rule $r'(p', q', v') \in P,$ $r' \neq r$}{
                \If{$v = v' \land ((p = p' \land q = q') \lor (p = q' \land q = p'))$}{
                    $M \leftarrow M \cup \{ r' \}$\;
                }
            }
            \If{$ \lvert M \rvert = 4$}{
                $I \leftarrow I \cup \{ $\invariant($M$)$ \}$\;
            }
        }
        $J \leftarrow \emptyset$\;
        \ForEach{trajectory $t \in$ T}{
            \ForEach{$predicate \ p \in \Psi, predicate \ q \in \Psi, variable \ v \in $ \var{$p_1$} $\cap$ \var{$p_2$}$, p \neq q$}{
                $J \leftarrow J \cup \{ $\invariant($p, q, v, s_{0, t}$)$ \}$\;
            }
        }
        $K \leftarrow $ \merge{$I, J$}\;
        \ForEach{ambiguous $\langle R_{a_{f}}^{+}, R_{a_{f}}^{-}, a_f \rangle$}{
            \If{$\exists k(p, q, v, op) \in K, \{p, q\} \subseteq R_{a_{f}}^{+} \cup R_{a_{f}}^{-}$}{
                \uIf{$op = \oplus \lor op = \odot$}{
                    Mark $p$ and $q$ as confirmed\;
                }
            }
        }
        \Return A\; 
    }
    \caption{Invariant Extraction}
    \label{fig:blackout-3}
\end{algorithm}

Blackout first tries to identify invariants in the action effects. 
It does so by looking for \textit{candidate primitive rules}, relations between predicates $p$ and $q$ asserting that $q$'s addition implies $p$'s addition (or removal, depending on whether $p$ and $q$ are in $\texttt{eff}^{+}$ or $\texttt{eff}^{-}$).
Candidate primitive rules are merged where possible to form \textit{proper primitive rules}. 
If these proper primitive rules cover the four cases for $p$ and $q$'s addition and removal,  they represent an invariant.

However, the invariants identified so far aren't necessarily true for the initial state of the trajectory. 
For instance, in the IPC problem files, the "wall" cells each get a \texttt{location} object with neither an \texttt{at} predicate nor a \texttt{clear} predicate attached. 
This breaks the mutual exclusivity invariant we observe when looking at just the action effects.
In addition, some predicates (namely \texttt{MOVE-DIR}, \texttt{IS-GOAL}, and \texttt{IS-NONGOAL}) never appear in the action effects at all, which means the above process is entirely blind to them, even though these predicates can encode useful information.
To solve both these issues, Blackout analyzes the initial state of the system $s_0$ when given a trajectory $t$.

Blackout includes a procedure $\texttt{Invariant}(p, q, v, s_0)$, which returns candidate invariant relations between $p$ and $q$ that hold in the initial state. 
These invariants are unioned with those from action effects to form the final list of invariants in the domain.
These invariants are guaranteed to hold in any state reachable from the initial state, provided that the effects discovered in step 1 are accurate.

\newcolumntype{L}[1]{>{\raggedright\let\newline\\\arraybackslash\hspace{0pt}}m{#1}}
\newcolumntype{C}[1]{>{\centering\let\newline\\\arraybackslash\hspace{0pt}}m{#1}}
\newcolumntype{R}[1]{>{\raggedleft\let\newline\\\arraybackslash\hspace{0pt}}m{#1}}

\begin{table}[h]
    \begin{tabular}{|C{0.2\columnwidth}|L{0.7\columnwidth}|} 
     \hline
     Invariant ($p R q)$ & \multicolumn{1}{|c|}{Meaning} \\
     \hline
     $\bot$ & Neither predicate can ever be present, and neither can ever be absent \\
     \hline
     $p \land q$ & Both predicates must always be present \\
     \hline
     $p \centernot\Longleftarrow q$ & First predicate is always present, second is always absent \\
     \hline
     $p$ & First predicate is always present, second may be present or absent \\
     \hline
     $p \centernot\implies q$ & First predicate is always absent, second is always present \\
     \hline
     $q$ & Second predicate is always present, first may be present or absent \\
     \hline
     $p \oplus q$ & Predicates are mutually exclusive \\
     \hline
     $p \lor q$ & At least one of the predicates must always be present \\
     \hline
     $p \downarrow q$ & Neither predicate can ever be present \\
     \hline
     $p \odot q$ & Predicates are equivalent: when one is present, the other is too \\
     \hline
     $\neg q$ & Second predicate is always absent, first can be present or absent \\
     \hline
     $p \Longleftarrow q$ & If the second predicate is present, the first must also be present \\
     \hline
     $\neg p$ & First predicate is always absent, second can be present or absent \\
     \hline
     $p \implies q$ & If the first predicate is present, the second must also be present \\
     \hline
     $p \uparrow q$ & At most one of the predicates can be present \\
     \hline
     $\top$ & Predicates are independent. Each can be present or absent, regardless of the other \\
     \hline
    \end{tabular}
    \caption{List of all possible invariants and their meanings}
    \label{tab:invariants}
\end{table}

These invariants are used to resolve ambiguous failed actions from step 2. 
If any invariant refers to two different predicates and their corresponding arguments in the same $R_{a_f}$, Blackout takes one of five possible actions on the $\texttt{pre}(a_f)$ sets based on the invariant type.
We document these in Table \ref{tab:inv-actions}.

\begin{table}[]
    \centering
    \begin{tabular}{c|c}
        Invariants & Action \\
        \hline
        $\bot, \land, \centernot\Longleftarrow, \centernot\Longrightarrow, \downarrow$ & Error \\
        $p, \lnot p$ & $q$ is confirmed as a precondition \\
        $q, \lnot q$ & $p$ is confirmed as a precondition \\
        $\lor. \Longrightarrow, \Longleftarrow, \uparrow, \top$ & No action \\
        $\oplus, \odot$ & $p$ and $q$ are confirmed as preconditions \\
    \end{tabular}
    \caption{The actions on the $R_{a_f}$ sets when a matching invariant is found}
    \label{tab:inv-actions}
\end{table}

\section{Evaluation}

\citeA{aineto2019104} introduced a novel metric for evaluating action models in the form of precision and recall, which we adapt to our task of producing a score which measures a player's understanding of a particular mechanic. 
This technique is a slight modification of their evaluation scheme.

We make the assumption that every action in the domain of a game defined in PDDL corresponds to a mechanic.
We compare the learned action model to the ground truth model and \textit{for each action} count the number of predicates in the learned model's preconditions and effects which are correct (true positives), extra (false positives) and missing (false negatives). 
We use this confusion matrix to compute the $F_1$-score for every action in the model and report it as the player's proficiency score for a particular mechanic given the model, as shown in Table \ref{tab:results}. 
We use the $F_1$-score since it is a meaningful way to combine precision and recall into a single number.

\begin{equation}
    F_1 = 2 \cdot \frac{\text{precision} \cdot \text{recall}}{\text{precision} + \text{recall}}
\end{equation}

\begin{table}[h]
    \centering
    \begin{tabular}{lcc}
        \multicolumn{1}{c}{Mechanic name} & \multicolumn{2}{c}{$F_1$-score} \\
                                 & $M_1$          & $M_2$          \\
        \texttt{move}                     & $0.267$        & $0.250$        \\
        \texttt{push-to-nongoal}          & $0.308$        & $0.216$        \\
        \texttt{push-to-goal}             & $0.222$        & $0.240$       
    \end{tabular}
    \caption{Proficiency scores for each mechanic} 
    \label{tab:results}
\end{table}

\begin{figure}
    \centering
    \includegraphics[width=\columnwidth]{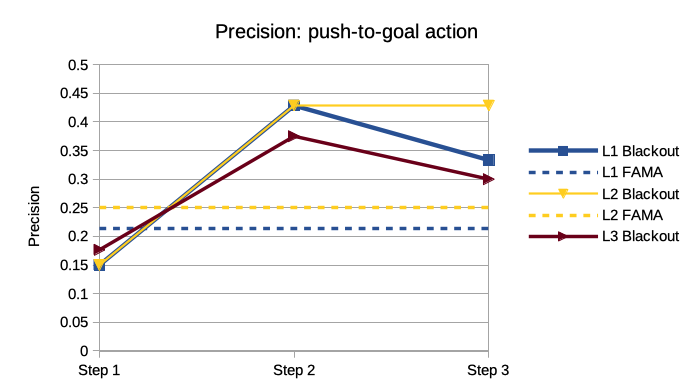}
    \caption{Precision scores for Blackout and FAMA for the push-to-goal action}
    \label{fig:precision-graph}
\end{figure}

\begin{table}[]
    \centering
    \begin{tabular}{c|c}
        Action & Recall \\
        \hline
        \texttt{move} & $0.2857$ \\
        \texttt{push-to-nongoal} & $0.3077$ \\
        \texttt{push-to-goal} & $0.2308$
    \end{tabular}
    \caption{Recall values for each action across all Blackout stages and FAMA, for levels $L_1$, $L_2$ and $L_3$}
    \label{tab:results-recall}
\end{table}

To compare FAMA and Blackout, we ran them both on traces generated manually from two hand-crafted levels ($L_1$ and $L_2$) and an instance from the IPC 2011 collection ($L_3$). 
The levels differ in complexity, with $L_1$ being the simplest and $L_3$ being the most complex. 
The traces also include failed actions.
We modify the evaluation scheme proposed by \citeA{aineto2019104} to measure precision and recall for \textit{each action} for each level. 
Since recall was found to be constant across all Blackout stages and FAMA, for all levels, differing only by actions, we report the recall values separately in Table \ref{tab:results-recall}.

Our measurements show that Blackout achieves better precision across all actions while achieving the same recall.

We do not report FAMA metrics for $L_3$ since we were unable to obtain an action model from due it the implementation running out of memory on our test machine.
We also do not report results for Blackout on |push-to-nongoal| since it doesn’t appear in the trajectory for that level and so Blackout does not know that it exists.
We display the precision scores for Blackout and FAMA for the |push-to-goal| action in Figure \ref{fig:precision-graph}. 
The other actions have similar trends, so we chose not to report them.

We notice that Step 3 of Blackout tends to decrease the precision scores in comparison to Step 2, as shown in Figure \ref{fig:precision-graph}. Whether this is indicative of a flaw in the algorithm or in the metric will be investigated in future work.

\subsection{Performance Scalability}

We use a machine with an Intel Core i7-9750H CPU with 16 GB RAM to compare the elapsed time and memory usage of FAMA and Blackout in learning action models. 
We attempt to learn a single action model for each of $L_1$, $L_2$ and $L_3$. 
To make our trajectories compatible with FAMA, we strip them of failed actions. 
We measure elapsed time and memory usage via \textit{wall clock time} and \textit{maximum resident set size} as output by the Unix \texttt{time} utility.
We repeat each run of the algorithm for each trajectory $30$ times and report the mean for each measured metric. 
Our findings for FAMA and for each step of Blackout are listed in Table \ref{tab:performance}.

\begin{table}[]
\centering
\begin{tabular}{cl|c|c|c|c|}
\cline{3-6}
\multirow{2}{*}{}                            &     & \multicolumn{3}{c|}{Blackout}     & \multirow{2}{*}{FAMA} \\ \cline{3-5}
                                             &     & $S_{1}$ & $S_{1+2}$ & $S_{1+2+3}$ &                       \\ \hline
\multicolumn{1}{|c|}{\multirow{2}{*}{$L_1$}} & Time (s) & $0.11$  & $0.10$    & $0.11$      & $10.59$               \\ \cline{2-6} 
\multicolumn{1}{|c|}{}                       & Mem. (MB) & $6.96$  & $6.90$    & $7.08$      & $1572.60$             \\ \hline
\multicolumn{1}{|c|}{\multirow{2}{*}{$L_2$}} & Time (s) & $0.10$  & $0.09$    & $0.11$      & $21.19$               \\ \cline{2-6} 
\multicolumn{1}{|c|}{}                       & Mem. (MB) & $6.80$  & $6.78$    & $6.99$      & $4632.52$             \\ \hline
\multicolumn{1}{|c|}{\multirow{2}{*}{$L_3$}} & Time (s) & $2.31$  & $2.34$    & $2.48$      & ---                   \\ \cline{2-6} 
\multicolumn{1}{|c|}{}                       & Mem. (MB) & $35.82$ & $35.78$   & $35.82$     & ---                   \\ \hline
\end{tabular}
\caption{Performance comparison between FAMA and Blackout for player modeling}
\label{tab:performance}
\end{table}

FAMA was unsuccessful in learning action models from a relatively large trajectory file ($\sim 1$ MB) due to running out of memory on the test machine; however, we expect player trajectories to be much larger in size ($> 1$ GB).

We see that Blackout to be much faster in learning action models than FAMA while having a smaller memory footprint.
However, further work needs to be done to assess performance in the case of more realistic trajectory sizes.

All of our code, including the tests, data, and Blackout implementation, is available in an open-source repository\footnote{\url{https://github.com/AbhijeetKrishnan/aml-for-player-modeling}}

\section{Conclusion and Future Work}

To the best of our knowledge, our work is the first application of AML to player modeling.
We demonstrated the feasibility of AML as a domain-agnostic player modeling approach by evaluating an existing AML algorithm for the task of player modeling, and we presented a technique to measure the player's mechanical proficiency using the learned model. We also introduced Blackout, a novel AML algorithm that incorporates action failures, compared the two approaches, and found that Blackout learns action models better and faster than FAMA. Based on these findings, we believe action model learning has the potential to serve as a useful player modeling technique in a wide range of games.

We were motivated in this project by a hypothesis that the cognitive process of mental model formation~\cite{wasserman2019bugs} could help design better AML algorithms for player modeling.
Blackout is informed by this motivation in updating action models based on failed actions, which are used similarly in mental model alignment~\cite{boyan2018model}. However, we have not yet conducted an experiment to measure how well our learned models match human players' mental models; we plan to do this in future work.

Our proposed measure of player competency will also benefit from refinement in future work.
Quantitatively estimating a player's mechanical proficiency leaves out higher-order skills learned by combining the use of mechanics that players need to learn in order to become proficient at a game \cite{cook_2007}. Future work will attempt to model and measure the acquisition of higher-order skills.

\bibliography{main}

\begin{thebibliography}{}

\bibitem[\protect\citeauthoryear{Aineto, Celorrio, and
  Onaindia}{2019}]{aineto2019104}
Aineto, D.; Celorrio, S.~J.; and Onaindia, E.
\newblock 2019.
\newblock Learning action models with minimal observability.
\newblock {\em Artificial Intelligence} 275:104 -- 137.

\bibitem[\protect\citeauthoryear{Biswas and Springett}{2018}]{biswas2018user}
Biswas, P., and Springett, M.
\newblock 2018.
\newblock User modeling.
\newblock {\em The Wiley Handbook of Human Computer Interaction Volume}  143.

\bibitem[\protect\citeauthoryear{Boyan, McGloin, and
  Wasserman}{2018}]{boyan2018model}
Boyan, A.; McGloin, R.; and Wasserman, J.~A.
\newblock 2018.
\newblock Model matching theory: A framework for examining the alignment
  between game mechanics and mental models.
\newblock {\em Media and Communication} 6(2):126--136.

\bibitem[\protect\citeauthoryear{Chakraborti \bgroup et al\mbox.\egroup
  }{2017}]{chakraborti2017cognitive}
Chakraborti, T.; Kambhampati, S.; Scheutz, M.; and Zhang, Y.
\newblock 2017.
\newblock {AI} challenges in human-robot cognitive teaming.
\newblock {\em CoRR} abs/1707.04775.

\bibitem[\protect\citeauthoryear{Chrysafiadi and
  Virvou}{2013}]{chrysafiadi2013student}
Chrysafiadi, K., and Virvou, M.
\newblock 2013.
\newblock Student modeling approaches: A literature review for the last decade.
\newblock {\em Expert Systems with Applications} 40(11):4715 -- 4729.

\bibitem[\protect\citeauthoryear{Coles \bgroup et al\mbox.\egroup
  }{2012}]{ipc2011survey}
Coles, A.; Coles, A.; Olaya, A.~G.; Jiménez, S.; López, C.~L.; Sanner, S.;
  and Yoon, S.
\newblock 2012.
\newblock A survey of the seventh international planning competition.
\newblock {\em AI Magazine} 33(1):83--88.
\newblock Copyright - Copyright Association for the Advancement of Artificial
  Intelligence Spring 2012; Document feature - ; Diagrams; Last updated -
  2018-10-06.

\bibitem[\protect\citeauthoryear{Cook}{2007}]{cook_2007}
Cook, D.
\newblock 2007.
\newblock The chemistry of game design.
\newblock Accessed May 15, 2020 from
  \url{https://www.gamasutra.com/view/feature/129948/the_chemistry_of_game_design.php}.

\bibitem[\protect\citeauthoryear{{Drachen}, {Canossa}, and
  {Yannakakis}}{2009}]{drachen2009}
{Drachen}, A.; {Canossa}, A.; and {Yannakakis}, G.~N.
\newblock 2009.
\newblock Player modeling using self-organization in tomb raider: Underworld.
\newblock In {\em 2009 IEEE Symposium on Computational Intelligence and Games},
   1--8.

\bibitem[\protect\citeauthoryear{Harrison and
  Roberts}{2011}]{harrison2011using}
Harrison, B., and Roberts, D.~L.
\newblock 2011.
\newblock Using sequential observations to model and predict player behavior.
\newblock In {\em Proceedings of the 6th International Conference on
  Foundations of Digital Games},  91--98.

\bibitem[\protect\citeauthoryear{Hooshyar, Yousefi, and
  Lim}{2018}]{hooshyar2018datadriven}
Hooshyar, D.; Yousefi, M.; and Lim, H.
\newblock 2018.
\newblock Data-driven approaches to game player modeling: A systematic
  literature review.
\newblock {\em ACM Comput. Surv.} 50(6).

\bibitem[\protect\citeauthoryear{Junghanns and
  Schaeffer}{1997}]{junghanns1997sokoban}
Junghanns, A., and Schaeffer, J.
\newblock 1997.
\newblock Sokoban: A challenging single-agent search problem.
\newblock In {\em In IJCAI Workshop on Using Games as an Experimental Testbed
  for AI Reasearch}.
\newblock Citeseer.

\bibitem[\protect\citeauthoryear{Machado, Fantini, and
  Chaimowicz}{2011}]{machado2011player}
Machado, M.~C.; Fantini, E.~P.; and Chaimowicz, L.
\newblock 2011.
\newblock Player modeling: Towards a common taxonomy.
\newblock In {\em 2011 16th international conference on computer games
  (CGAMES)},  50--57.
\newblock IEEE.

\bibitem[\protect\citeauthoryear{Nogueira \bgroup et al\mbox.\egroup
  }{2014}]{nogueira2014fuzzy}
Nogueira, P.~A.; Aguiar, R.; Rodrigues, R.~A.; Oliveira, E.~C.; and Nacke, L.
\newblock 2014.
\newblock Fuzzy affective player models: A physiology-based hierarchical
  clustering method.
\newblock In {\em Tenth Artificial Intelligence and Interactive Digital
  Entertainment Conference}.

\bibitem[\protect\citeauthoryear{{Pedersen}, {Togelius}, and
  {Yannakakis}}{2010}]{pedersen2010content}
{Pedersen}, C.; {Togelius}, J.; and {Yannakakis}, G.~N.
\newblock 2010.
\newblock Modeling player experience for content creation.
\newblock {\em IEEE Transactions on Computational Intelligence and AI in Games}
  2(1):54--67.

\bibitem[\protect\citeauthoryear{Pfau, Smeddinck, and
  Malaka}{2018}]{pfau2018dpbm}
Pfau, J.; Smeddinck, J.~D.; and Malaka, R.
\newblock 2018.
\newblock Towards deep player behavior models in mmorpgs.
\newblock In {\em Proceedings of the 2018 Annual Symposium on Computer-Human
  Interaction in Play}, CHI PLAY ’18,  381–392.
\newblock New York, NY, USA: Association for Computing Machinery.

\bibitem[\protect\citeauthoryear{Serafini and
  Traverso}{2019}]{serafini2019incremental}
Serafini, L., and Traverso, P.
\newblock 2019.
\newblock Incremental learning of discrete planning domains from continuous
  perceptions.
\newblock {\em arXiv preprint arXiv:1903.05937}.

\bibitem[\protect\citeauthoryear{Smith \bgroup et al\mbox.\egroup
  }{2011}]{smith2011inclusive}
Smith, A.~M.; Lewis, C.; Hullet, K.; Smith, G.; and Sullivan, A.
\newblock 2011.
\newblock An inclusive view of player modeling.
\newblock In {\em Proceedings of the 6th International Conference on
  Foundations of Digital Games},  301--303.

\bibitem[\protect\citeauthoryear{Snodgrass, Mohaddesi, and
  Harteveld}{2019}]{snodgrass2019peas}
Snodgrass, S.; Mohaddesi, O.; and Harteveld, C.
\newblock 2019.
\newblock Towards a generalized player model through the peas framework.
\newblock In {\em Proceedings of the 14th International Conference on the
  Foundations of Digital Games}, FDG ’19.
\newblock New York, NY, USA: Association for Computing Machinery.

\bibitem[\protect\citeauthoryear{Summerville \bgroup et al\mbox.\egroup
  }{2016}]{summerville2016learning}
Summerville, A.; Guzdial, M.; Mateas, M.; and Riedl, M.~O.
\newblock 2016.
\newblock Learning player tailored content from observation: Platformer level
  generation from video traces using lstms.
\newblock In {\em Twelfth Artificial Intelligence and Interactive Digital
  Entertainment Conference}.

\bibitem[\protect\citeauthoryear{Wang \bgroup et al\mbox.\egroup
  }{2018}]{pengcheng2018fidelity}
Wang, P.; Rowe, J.; Min, W.; Mott, B.; and Lester, J.
\newblock 2018.
\newblock High-fidelity simulated players for interactive narrative planning.
\newblock In {\em Proceedings of the 27th International Joint Conference on
  Artificial Intelligence}, IJCAI’18,  3884–3890.
\newblock AAAI Press.

\bibitem[\protect\citeauthoryear{Wang}{1995}]{wang1995learning}
Wang, X.
\newblock 1995.
\newblock Learning by observation and practice: An incremental approach for
  planning operator acquisition.
\newblock In {\em Machine Learning Proceedings 1995}. Elsevier.
\newblock  549--557.

\bibitem[\protect\citeauthoryear{Wasserman and Koban}{2019}]{wasserman2019bugs}
Wasserman, J.~A., and Koban, K.
\newblock 2019.
\newblock Bugs on the brain: A mental model matching approach to cognitive
  skill acquisition in a strategy game.
\newblock {\em Journal of Expertise/June} 2(2).

\bibitem[\protect\citeauthoryear{Yannakakis and
  Maragoudakis}{2005}]{yannakakis2005}
Yannakakis, G.~N., and Maragoudakis, M.
\newblock 2005.
\newblock Player modeling impact on player's entertainment in computer games.
\newblock In Ardissono, L.; Brna, P.; and Mitrovic, A., eds., {\em User
  Modeling 2005},  74--78.
\newblock Berlin, Heidelberg: Springer Berlin Heidelberg.

\bibitem[\protect\citeauthoryear{Yannakakis \bgroup et al\mbox.\egroup
  }{2013}]{yannakakis2013player}
Yannakakis, G.~N.; Spronck, P.; Loiacono, D.; and Andr{\'e}, E.
\newblock 2013.
\newblock Player modeling.

\end{thebibliography}
\bibliographystyle{aaai}

\end{document}